\pdfoutput=1

\documentclass[11pt]{article}

\usepackage[final]{acl}

\usepackage{times}
\usepackage{latexsym}
\usepackage{graphicx}
\usepackage{booktabs}
\usepackage{xspace}
\usepackage[T1]{fontenc}

\usepackage[utf8]{inputenc}

\usepackage{microtype}

\usepackage{inconsolata}
\usepackage{hyperref}
\title{iREL at SemEval-2024 Task 9: Improving Conventional Prompting Methods for Brain Teasers}

\author{Harshit Gupta, Manav Chaudhary, Tathagata Raha, \\ \bf Shivansh Subramanian and Vasudeva Varma\\
International Institute of Information Technology, Hyderabad (IIIT-H) \\
\{harshit.g, manav.chaudhary, tathagata.raha, shivansh.s\}@research.iiit.ac.in \\
vv@iiit.ac.in}

\begin{document}
\maketitle
\begin{abstract}
This paper describes our approach for SemEval-2024 Task 9: BRAINTEASER: A Novel Task Defying Common Sense. The BRAINTEASER task comprises multiple-choice Question Answering designed to evaluate the models' lateral thinking capabilities. It consists of Sentence Puzzle and Word Puzzle subtasks that require models to defy default common-sense associations and exhibit unconventional thinking. We propose a unique strategy to improve the performance of pre-trained language models, notably the Gemini 1.0 Pro Model, in both subtasks. We employ static and dynamic few-shot prompting techniques and introduce a model-generated reasoning strategy that utilizes the LLM’s reasoning capabilities to improve performance. Our approach demonstrated significant improvements, showing that it performed better than the baseline models by a considerable margin but fell short of performing as well as the human annotators, thus highlighting the efficacy of the proposed strategies.
We have made our code open-sourced for the replicability of our methods. \footnote{\url{https://github.com/TheAthleticCoder/iREL-at-SemEval-2024-Task-9.git}}
\end{abstract}

\section{Introduction}
Human cognition is characterized by two distinct modes of thinking: vertical and lateral \citep{waks1997lateral}. Vertical thinking, often called logical or convergent thinking, follows a structured analytical process based on reasoning and established rules. By contrast, lateral thinking, or "thinking outside the box," is a creative and divergent process that challenges conventional assumptions and explores novel perspectives. 

Vertical thinking and lateral thinking are both complementary \citep{dingli2008thinking}. Vertical thinking is known for its selectivity and focus, while lateral thinking is known for its creativity and ability to generate alternative approaches and perspectives. It is crucial to recognize the value of both types of thinking and utilize them in a balanced manner to achieve optimal results.

We work on two subtasks (Sentence Puzzle and Word Puzzle) introduced as part of the SemEval-2023 Task 9: BRAINTEASER: A Novel Task Defying Common Sense \citep{jiang-ilievski-ma:2024:SemEval2024}. The Sentence-type brain teasers contain puzzle-defying common sense teasers centred around sentence snippets. For instance, the question "What has a bed but no head, a mouth but no teeth?" challenges the default association of beds with people and forces the solver to consider other possibilities, such as a river (which has a riverbed but no heads or teeth).
In Word-type brain teasers, the answer violates the default meaning of the word and focuses on the letter composition of the target question. For example, the question "What word becomes shorter when you add two letters to it?" challenges the assumption that adding letters to a word would make it longer and requires the solver to recognize that the word "short" becomes "shorter" when "er" is added to it.

We used the Gemini 1.0 Pro Model \citep{team2023gemini} to evaluate the model's performance in zero-shot and few-shot settings. We also make notable enhancements in the few-shot setting. Firstly, by employing contextualized question selection, we ensure that the model is exposed to more relevant examples by identifying questions from the training set that closely resemble those in the test set. Secondly, we enable the model to generate explanations for correct answer choices during training through reason generation, thereby deepening its comprehension of the examples. These approaches have demonstrated improvements in the evaluation scores.


\section{Related Work}
\citet{inbook}, \citet{hendrycks2021measuring}, and \citet{singhal2023expertlevel} demonstrate recent advancements in multiple-choice question answering. They achieve this by developing new datasets and evaluating large language models (LLMs) on them, thus contributing significantly to the field's progress.

\citet{xie2023olagpt} presents OlaGPT, an innovative framework designed to enhance the reasoning capabilities of large language models (LLMs) by drawing inspiration from human cognitive architecture. OlaGPT integrates cognitive modules such as attention, memory, reasoning, and learning, emphasizing a reasoning module that simulates human-like thought processes. The module then enables OlaGPT to create multiple agents and utilize various thinking templates, including lateral and integrative thinking, to solve reasoning problems effectively. 

\citet{huang2023lateval} introduces a novel evaluation benchmark to assess a model's lateral thinking abilities in an interactive framework, utilizing Lateral Thinking Puzzles as the context. 

\citet{meng2024divide} proposes a divide-and-conquer approach to LLM reasoning. It involves categorizing questions into subsets based on statistical confidence scores (CS), followed by targeted interventions such as Prior Knowledge-based Reasoning (PKR) and Filter Choices-based Reasoning (FCR) to address nuanced and demanding tasks.

\section{Data}
\label{sec:data}
The primary dataset \citep{jiang-etal-2023-brainteaser} used in this study encompasses data pertinent to two subtasks: sentence puzzles and word puzzles.
The puzzle is presented in a single correct MCQ format, where each puzzle consists of a question and several options. Among these options, only one is the correct answer to the puzzle.
Creating multiple-choice questions challenges balancing fairness and intellectual engagement \citep{ma2021knowledge}. This necessitates carefully curating distractors that are not only incorrect but also sufficiently challenging. It is worth noting that within the training data for both puzzles, every brainteaser was accompanied by two distractor options alongside the correct option. Please refer to Table \ref{tab:dataset_subsets} for specific sample numbers.

\begin{table}[htbp]
\centering
\caption{Dataset Details}
\label{tab:dataset_subsets}
\begin{tabular}{@{}lcc@{}}
\toprule 
\hline
\textbf{Type of Puzzle} & \textbf{Train Samples} & \textbf{Test Samples} \\ \hline
\midrule
Sentence Puzzle & 507 & 120 \\
Word Puzzle & 396 & 96 \\ \hline
\bottomrule
\end{tabular}
\end{table}

\section{Methodology}
\subsection{Zero-Shot Prompting} 
Initially, we conducted experiments using a zero-shot approach\citep{brown2020language}. In this method, we presented the model with questions and their multiple-choice options and asked it to identify the correct option. Subsequently, we improved the zero-shot approach by introducing the model to a sentence or word puzzle concept depending on the specific subtask under evaluation. We provided the model with the puzzle definition and then asked it to select the correct option from the choices. This modified zero-shot prompt template can be found in App.\ref{app:a}.

\subsection{Few-Shot Prompting} 
To enhance the model's performance, we implemented a few-shot prompting technique \citep{brown2020language}. This methodology involved presenting a variable number of examples to the model to facilitate in-context learning for the lateral thinking task. Subsequently, the model was prompted to identify the correct option among the provided choices. App.\ref{app:b} shows examples of few-shot prompt templates.

\subsection{Contextualized Example Selection}
The method outlined above relies on a fixed set of examples for model prompting. We modify this by applying a Dynamic Few-Shot prompting approach inspired by \citet{nori2023generalist}, which enables in-context learning. This method involves selecting samples from the train data that closely match the semantic content of the samples posed in the test data. Initially, all questions from both the training and testing datasets undergo encoding using BERT-Large \citep{devlin2019bert}, a pre-trained transformer-based language model. Subsequently, Cosine Similarity is utilized to calculate similarity scores between each question in the testing dataset and all questions in the training dataset.

Based on these similarity scores, we select the $top-n$ most similar examples from the training dataset for each question in the testing dataset, where $n$ varies based on the desired number of examples to be used in the prompt. This dynamic selection process aims to leverage more relevant examples from the training data to allow for better in-context learning while evaluating each test sample.

\subsection{Self-Generated Reasoning}
In Section \ref{sec:data} of this paper, we discuss how our training data contains the correct option and two distractor options for each question. We utilize this information to prompt the Gemini Model\citep{team2023gemini} and GPT-4 Model \citep{achiam2023gpt} to produce detailed reasoning about why the correct option is correct and why the distractor options are incorrect, highlighting potential confusion for test-takers.
We include the models' reasoning and examples from the training data during inference on the testing data. This approach aims to improve the quality and precision of the models by providing detailed insights into the reasoning behind the options making up the example. The prompt template used to generate the reasons and some examples of model-generated reasons for the samples from the training data are provided in App.\ref{app:c},\ref{app:d}.

\subsection{Model and Hyperparameters} The Gemini Pro 1 Model \citep{team2023gemini} was used as the primary model in this study. The temperature parameter\citep{brown2020language} was set to $0.1$, to guide the model to indicate its belief regarding the correct option based solely on its pre-existing knowledge base. This low-temperature setting was chosen to minimize the generation of creative or unexpected outputs.
Additionally, we set both the \textit{$top\_p$} and \textit{$top\_k$} parameters \citep{brown2020language} to $1$. By restricting the model in this manner, we aimed to maintain the relevance and coherence of the responses within the context of our research tasks.

\begin{table*}[htbp]
\centering
\caption{Results of the Sentence Puzzle subtask: Ori = Original, Sem = Semantic, Con = Context, SE = Static Examples, DE = Dynamic Examples, GPTR = GPT4 Reasoning}
\label{tab:sentence_puzzle_scores}
\begin{tabular}{lcccccc}
\hline
\textbf{Strategy} & \textbf{Ori} & \textbf{Sem} & \textbf{Con} & \textbf{Ori \& Sem} & \textbf{Ori \& Sem \& Con} & \textbf{Overall} \\ \hline
\multicolumn{7}{c}{\textbf{Baseline}} \\
\hline
Chat-GPT 0 shot & 0.6077 & 0.5933 & 0.6794 & 0.5072 & 0.3971 & 0.6268 \\
Roberta-L & 0.4354 & 0.4019 & 0.4641 & 0.3301 & 0.2010 & 0.4338 \\
Human & 0.9074 & 0.9074 & 0.9444 & 0.9074 & 0.8889 & 0.9198 \\
\hline
\multicolumn{7}{c}{\textbf{Zero-Shot}} \\
\hline
Direct Prompt & 0.775 & 0.725 & 0.575 & 0.700 & 0.525 & 0.692 \\
Definition Prompt & 0.775 & 0.725 & 0.700 & 0.700 & 0.575 & 0.733 \\
\hline
\multicolumn{7}{c}{\textbf{Few-Shot}} \\
\hline
1 Shot + SE + Reason & \textbf{0.800} & \textbf{0.750} & 0.675 & 0.725 & 0.575 & 0.742 \\
3 Shot + SE + Reason & 0.775 & 0.700 & 0.750 & 0.700 & 0.625 & 0.742 \\
5 Shot + SE + Reason & \textbf{0.800} & 0.700 & 0.750 & 0.700 & \textbf{0.650} & 0.750 \\
1 Shot + SE + GPTR & \textbf{0.800} & \textbf{0.750} & \textbf{0.775} & \textbf{0.750} & \textbf{0.650} & \textbf{0.775} \\
3 Shot + SE + GPTR & \textbf{0.800} & \textbf{0.750} & 0.725 & 0.725 & 0.600 & 0.758 \\
5 Shot + SE + GPTR & \textbf{0.800} & 0.725 & 0.750 & 0.725 & 0.600 & 0.758 \\
1 Shot + DE & 0.750 & 0.700 & 0.650 & 0.675 & 0.550 & 0.700 \\
3 Shot + DE & 0.775 & 0.700 & 0.750 & 0.675 & 0.600 & 0.742 \\
5 Shot + DE & 0.775 & 0.725 & 0.750 & 0.700 & 0.625 & 0.750 \\
1 Shot + DE + Reason & 0.775 & 0.700 & 0.675 & 0.700 & 0.575 & 0.717 \\
3 Shot + DE + Reason & \textbf{0.800} & 0.725 & 0.750 & 0.700 & 0.625 & 0.758 \\
5 Shot + DE + Reason & \textbf{0.800} & 0.725 & 0.750 & 0.725 & \textbf{0.650} & 0.758 \\
\hline
\end{tabular}
\end{table*}

\begin{table*}[htbp]
\centering
\caption{Results of the Word Puzzle subtask: Ori = Original, Sem = Semantic, Con = Context, SE = Static Examples, DE = Dynamic Examples, GPTR = GPT4 Reasoning}
\label{tab:word_puzzle_scores}
\begin{tabular}{lcccccc}
\hline
\textbf{Strategy} & \textbf{Ori} & \textbf{Sem} & \textbf{Con} & \textbf{Ori \& Sem} & \textbf{Ori \& Sem \& Con} & \textbf{Overall} \\ \hline
\multicolumn{7}{c}{\textbf{Baseline}} \\
\hline
Chat-GPT 0 shot & 0.5610 & 0.5244 & 0.5183 & 0.4390 & 0.2927 & 0.5346 \\
Roberta-L & 0.1951 & 0.1951 & 0.2317 & 0.1463 & 0.061 & 0.2073 \\
Human & 0.9167 & 0.9167 & 0.9167 & 0.9167 & 0.8958 & 0.9167 \\
\hline
\multicolumn{7}{c}{\textbf{Zero-Shot}} \\
\hline
Direct Prompt & 0.688 & 0.438 & 0.562 & 0.375 & 0.281 & 0.562 \\
Definition Prompt & 0.719 & 0.719 & 0.781 & 0.562 & 0.531 & 0.740 \\
\hline
\multicolumn{7}{c}{\textbf{Few-Shot}} \\
\hline
1 Shot + SE + Reason & 0.781 & 0.688 & \textbf{0.844} & 0.562 & 0.562 & 0.771 \\
3 Shot + SE + Reason & 0.812 & 0.656 & \textbf{0.844} & 0.562 & 0.531 & 0.771 \\
5 Shot + SE + Reason & 0.719 & 0.719 & \textbf{0.844} & 0.562 & 0.531 & 0.760 \\
1 Shot + SE + GPTR & 0.562 & 0.531 & 0.406 & 0.500 & 0.250 & 0.500 \\
3 Shot + SE + GPTR & 0.781 & 0.625 & 0.812 & 0.562 & 0.500 & 0.740 \\
5 Shot + SE + GPTR & 0.750 & 0.719 & \textbf{0.844} & 0.625 & 0.562 & 0.771 \\
1 Shot + DE & 0.844 & 0.688 & 0.719 & 0.656 & 0.562 & 0.750 \\
3 Shot + DE & 0.844 & 0.719 & 0.781 & 0.656 & 0.625 & 0.781 \\
5 Shot + DE & \textbf{0.875} & 0.781 & 0.812 & \textbf{0.750} & \textbf{0.656} & \textbf{0.823} \\
1 Shot + DE + Reason & 0.844 & 0.688 & 0.812 & 0.656 & 0.625 & 0.781 \\
3 Shot + DE + Reason & 0.750 & \textbf{0.812} & 0.750 & 0.656 & 0.594 & 0.771 \\
5 Shot + DE + Reason & 0.750 & \textbf{0.812} & 0.812 & 0.688 & 0.594 & 0.792 \\
\hline
\end{tabular}
\end{table*}

\begin{figure}[htbp]
    \centering
    \includegraphics[width=0.9\linewidth]{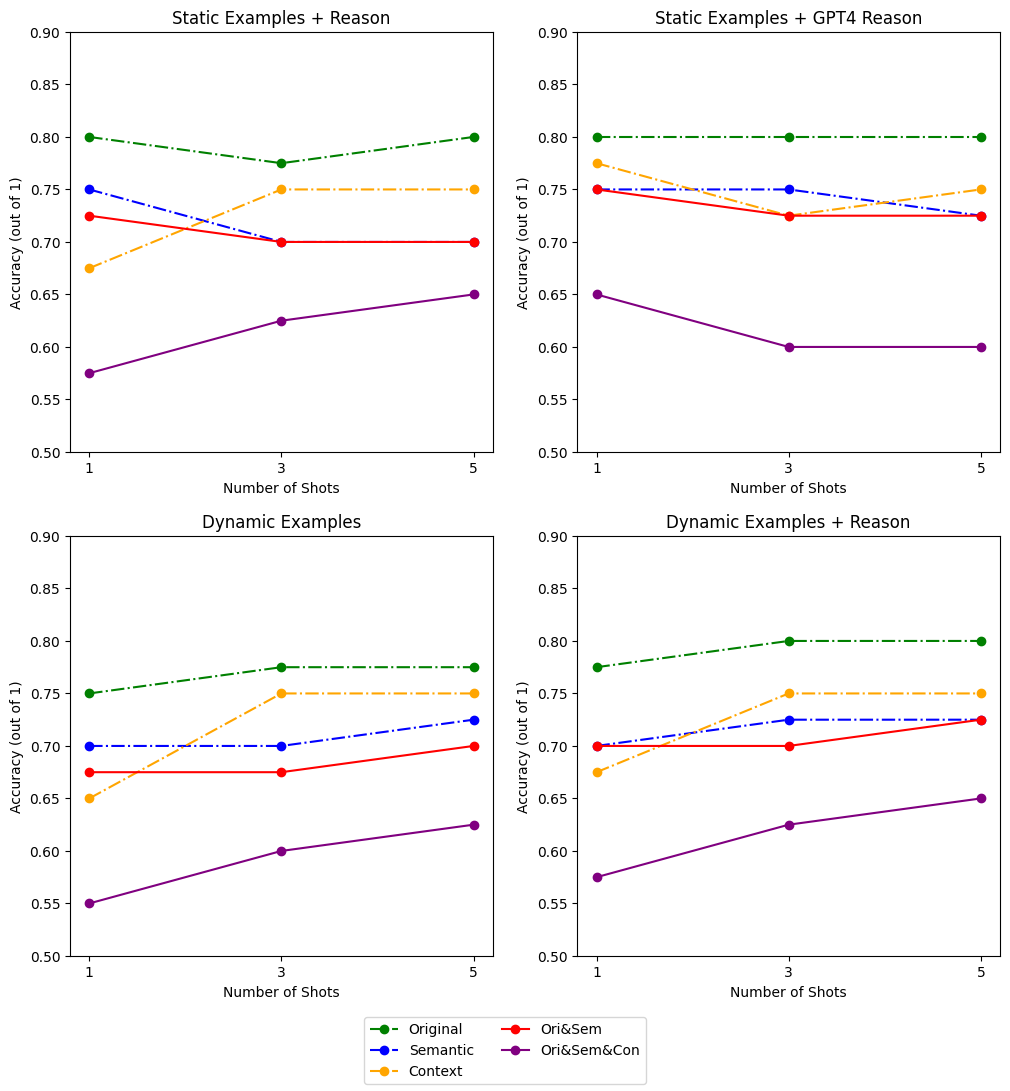} 
    \caption{Few-shot prompting performance on the Sentence Puzzle subtask}
    \label{fig:sentpuzz}
\end{figure}

\begin{figure}[htbp]
    \centering
    \includegraphics[width=0.9\linewidth]{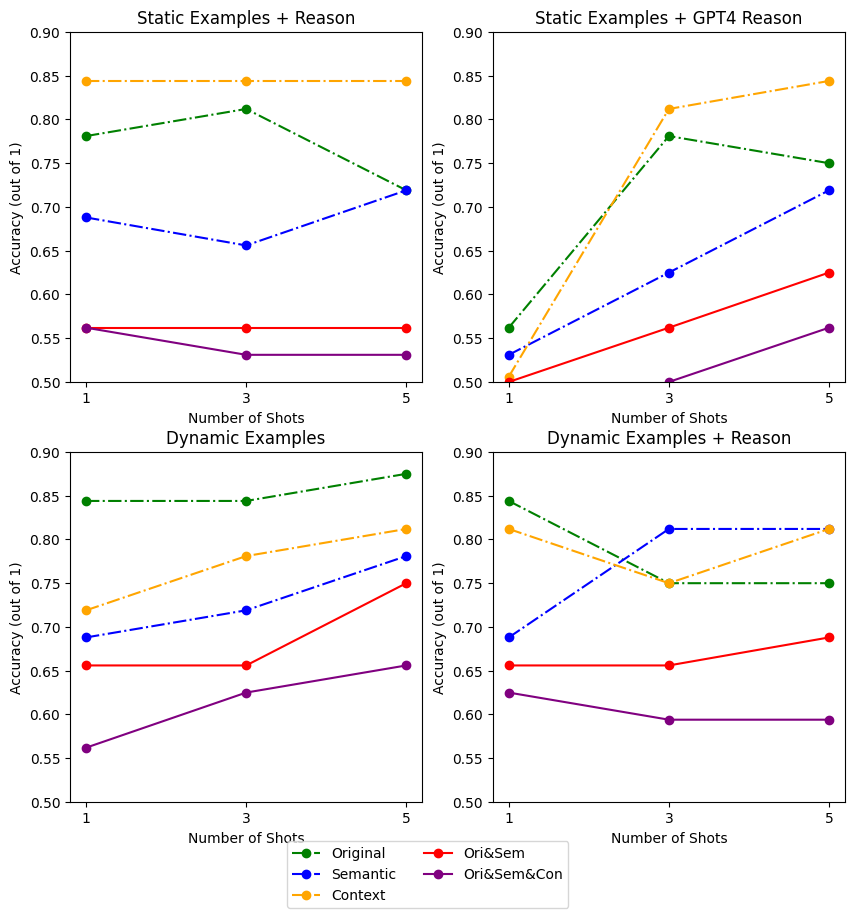} 
    \caption{Few-shot prompting performance on the Word Puzzle subtask}
    \label{fig:wordpuzz}
\end{figure}

\section{Results}
The results of our experiments and ablation studies conducted for the Sentence Puzzle and Word Puzzle subtasks are presented in Tables \ref{tab:sentence_puzzle_scores} and \ref{tab:word_puzzle_scores}, respectively. The baseline scores from the best-performing model in the Instruction, Commonsense, and Human categories, as provided by \citet{jiang2023brainteaser}, serve as benchmarks for comparison. Notably, across all six columns of scores, the Zero-Shot approaches and Few-Shot approaches outperform the baseline Chat-GPT 0 Shot and Roberta-L models by a significant margin.

Employing the zero-shot approach of providing the model with the definition of the sentence or word puzzle (Definition Prompting) yields superior performance across both subtasks compared to simply prompting it to indicate the correct option (Direct Prompting). The significant improvement contributing factor to the overall score is evident in the significant improvements observed in the context reconstruction scores(Con).

Based on the results shown in Figure \ref{fig:sentpuzz} and Table \ref{tab:sentence_puzzle_scores}, it appears that incorporating GPT-4's reasoning alongside static examples (SE) proved to be the most effective strategy for tackling Sentence Puzzles. The tested strategies also demonstrated improved outcomes when more examples were employed, emphasizing the crucial role of using more in-context examples. However, it is noteworthy that adding self-generated reasoning to the few-shot strategy with dynamic examples did not yield a commensurate improvement; instead, it resulted in a trade-off. While it enhanced the scores for semantically related questions, it came at the expense of the performance on other question types.

Contrary to expectations, the findings in Figure \ref{fig:wordpuzz} and Table \ref{tab:word_puzzle_scores} reveal that static examples and reasoning work just as well and even better than dynamic examples and reasoning in the few-shot learning context for the Word Puzzle task. Specifically, incorporating self-generated reasoning alongside dynamic examples led to no significant advancements, indicating that the presumed benefits of dynamic examples did not materialize as expected.

\section{Conclusion}
In our work, our extensive experimentation demonstrates that the Gemini Pro 1 Model can perform lateral thinking tasks and demonstrate significant improvements in both Sentence Puzzle and Word Puzzle subtasks by employing static and dynamic example selection coupled with self-generated reasoning strategies. We achieved notable enhancements over the baseline models and observed minor improvements as we increased the number of examples used for prompting. While our approach demonstrates notable progress, it still falls short of the performance of human annotators, indicating that further research and development are necessary to bridge this gap. 

\bibliography{custom}

\appendix
\section{Zero-Shot Prompt Template}
\label{app:a}
Here, we provide the modified zero-shot prompt templates for the sentence and word puzzles.
\subsection{Sentence Puzzle}
\texttt{Welcome to the sentence-play puzzle challenge! You are presented with a question based on a sentence-play puzzle. It means that the question is a sentence-type brain teaser where the puzzle-defying commonsense is centered on sentence snippets. Remember to pay attention to the details mentioned and indicate the option number you believe is correct for the question: \\
Question: \{question\} \\
Choices: \{choices\}}

\subsection{Word Puzzle}
\texttt{Welcome to the word-play puzzle challenge! You are presented with a question based on a word-play puzzle. It means that the question is a brain teaser where the answer violates the default meaning of the word and focuses on the letter composition of the target question. Remember to pay attention to the details mentioned and indicate the option number you believe is correct for the question: \\
Question: \{question\} \\
Choices: \{choices\}}

\section{Few-Shot Prompt Template}
\label{app:b}
We provide the 2-shot prompt templates for both puzzles here as an example of the few-shot prompting approach. 

\subsection{Sentence Puzzle}
\texttt{Welcome to the sentence-play puzzle challenge! Here, you will be presented with a question based on a sentence-play puzzle. It means that the question is a sentence-type brain teaser where the puzzle-defying commonsense is centred on sentence snippets.\\
We have given you two examples below to help you understand the puzzle challenge better. \\
Example 1: \\
Question: \{question\} \\
Choices:\{choices\} \\
Correct Option: \{correct choice\} \\
Example 2: \\
Question: \{question\} \\
Choices: \{choices\} \\
Correct Option: \{correct choice\} \\
Now, we shall be giving you the puzzle you need to solve. Remember to pay attention to the details mentioned and indicate the option number you believe is correct for the question: \\
Question: \{question\} \\
Choices: \{choices\}}

\subsection{Word Puzzle}
\texttt{Welcome to the word-play puzzle challenge! Here, you will be presented with a question based on a word-play puzzle. It means that the question is a brain teaser where the answer violates the default meaning of the word and focuses on the letter composition of the target question. \\
We have given you two examples below to help you understand the puzzle challenge better.\\
Example 1: \\
Question: \{question\} \\
Choices: \{choices\} \\
Correct Option: \{correct choice\} \\
Example 2: \\
Question: \{question\} \\
Choices: \{choices\} \\
Correct Option: \{correct choice\} \\
Now, we shall be giving you the puzzle you need to solve. Remember to pay attention to the details mentioned and indicate the option number you believe is correct for the question: \\
Question: \{question\} \\
Choices: \{choices\}}

\section{Prompt Template for Self-Generated Reasoning}
\label{app:c}
We provide the self-generated reasoning prompt template for the sentence and word puzzles. 
\subsection{Sentence Puzzle}
\texttt{\#\#\# CONTEXT \\
We are presented with a question based on a sentence-play puzzle. It means that the question is a sentence-type brain teaser where the puzzle-defying commonsense is centered on sentence snippets. \\
\#\#\# OBJECTIVE \\
We have provided you below with the question, the answer choices and the correct option number and choice. We have also provided you with what the distractor choice was. This distractor was aimed to throw you off the correct answer.
You need to provide the reasoning for why the option is correct. \\
Question: \{question\} \\
Choices: \{choices\} \\
Correct Option: Option \{option number\} : \{relevant content\} \\
Distractor Choice: \{distractor option content\} \\
\#\#\# RESPONSE \\
Provide the reasoning for why the given correct option is correct and what should be taken care of so that the distractor choice is not chosen. \\
\#\#\# REASONING}

\subsection{Word Puzzle}
\texttt{\#\#\# CONTEXT \\
We are presented with a question based on a word-play puzzle. It means that the question is a brain teaser where the answer violates the default meaning of the word and focuses on the letter composition of the target question.  \\
\#\#\# OBJECTIVE \\
We have provided you below with the question, the answer choices and the correct option number and choice. We have also provided you with what the distractor choice was. This distractor was aimed to throw you off the correct answer.
You need to provide the reasoning for why the option is correct. \\
Question: \{question\} \\
Choices: \{choices\} \\
Correct Option: Option \{option number\} : \{relevant content\} \\
Distractor Choice: \{distractor option content\} \\
\#\#\# RESPONSE \\
Provide the reasoning for why the given correct option is correct and what should be taken care of so that the distractor choice is not chosen. \\
\#\#\# REASONING}

\section{Examples of Self-Generated Reasoning}
\label{app:d}
We have provided examples of the reasoning generated by the Gemini Pro 1 and GPT-4 Models over the training data. 

\subsection{Reasoning using Gemini Pro 1}
\subsubsection{Example 1}
\texttt{\textbf{Question:} There was once a bad wizard. He abducted three women from their homes and transformed them into identical rose plants. They were placed in his garden. One of the women pleaded with the magician to allow her to see her husband and kids. He concurred. He took the woman to her home at night. He arrived early in the morning and drove her home. The spouse made the decision to go get her one day. He entered the wizard's garden as a result. He kept glancing at the three identical rose bushes, trying to determine which one might be his wife. He took his wife home as soon as he realized the solution. How did he recognize his wife among the rosebushes? \\
\textbf{Choices:} \\
Option 1: The husband knew which rose bush was his wife because it was the only one that still had her distinctive scent. \\
Option 2: The wizard brought the rosebush to her home at night and returned her to the garden in the morning. Therefore, she was the only plant without dew.\\
Option 3: The husband knew which rose bush was his wife because it was the only one that leaned towards him as he approached.\\
Option 4: None of above.\\
\textbf{Correct Option: 2} \\
\textbf{Reason for Correct Option:} Reasoning for the Correct Option: \\
Option 2: "The wizard brought the rosebush to her home at night and returned her to the garden in the morning. Therefore, she was the only plant without dew." is correct because: \\
- The passage states that the wizard took the woman to her home at night and returned her to the garden in the morning. \\
- This means that the rosebush that was his wife would be the only one without dew on it in the morning, as it had been indoors all night. \\
- The other rosebushes would have dew on them because they had been outside all night. \\
Distractor Choice Analysis: \\
Option 1: "The husband knew which rose bush was his wife because it was the only one that still had her distinctive scent." is a distractor choice because: \\
- The passage does not mention anything about the women retaining their distinctive scents after being transformed into rosebushes. \\
- Therefore, it is not possible for the husband to identify his wife by her scent. \\
Conclusion: \\
Therefore, the correct option is Option 2: "The wizard brought the rosebush to her home at night and returned her to the garden in the morning. Therefore, she was the only plant without dew." because it is the only option that is supported by the information in the passage.
}

\subsubsection{Example 2}
\texttt{\textbf{Question:} What is the most shocking city? \\
\textbf{Choices:} \\
Option 1: Urban city. \\
Option 2: Inner city.\\
Option 3: Electricity.\\
Option 4: None of above.\\
\textbf{Correct Option: 3} \\
\textbf{Reason for Correct Option:} 
Reasoning: \\
The question is "What is the most shocking city?".
The correct answer is "Electricity".
Electricity is a form of energy that can cause shocks.
A city is a large urban area.
Therefore, the most shocking city is Electricity. \\
Distractor Choice: \\
The distractor choice is "Inner city".
Inner city is a part of a city that is typically characterized by poverty and crime.
Inner city is not a form of energy that can cause shocks.
Therefore, inner city is not the most shocking city. \\
To avoid choosing the distractor choice, one should: \\
- Pay attention to the question and the answer choices. \\
- Understand the meaning of the words in the question and the answer choices. \\
- Use logic and reasoning to eliminate the incorrect answer choices. \\
}

\subsection{Reasoning using GPT-4}
\subsubsection{Example 1}
\texttt{\textbf{Question:} An electric train is going south at 98 mph. The wind is blowing northeast. Which direction is the smoke blowing? \\
\textbf{Choices:} \\
Option 1: Northeast. \\
Option 2: West.\\
Option 3: South.\\
Option 4: None of above.\\
\textbf{Correct Option:} 4\\
\textbf{Reason for Correct Option:} To solve this sentence-play puzzle, let's analyze the key elements of the question:\\
Electric Train: This is a crucial piece of information. Electric trains do not produce smoke as they do not burn fuel in the same way that steam or diesel trains do.\\
Direction of Train: The train is going south. This would only be relevant if we were dealing with a train that produces smoke.\\
Wind Direction: The wind is blowing northeast. Normally, this would affect the direction of the smoke if the train produced any.\\
Given these points, the key detail here is that the train is electric. Therefore, the direction of the smoke is a trick question because there would be no smoke produced by an electric train.
}

\subsubsection{Example 2}
\texttt{\textbf{Question:} What kind of lamp emits no light? \\
\textbf{Choices:} \\
Option 1: Oil lamp.\\
Option 2: LED lamp.\\
Option 3: Clamp.\\
Option 4: None of above.\\
\textbf{Correct Option:} 3 \\
\textbf{Reason for the Correct Option:} This puzzle requires us to think beyond the literal uses of the words provided, focusing on the play on words involved in the question and the choices given. The question asks, "What kind of lamp emits no light?" Here are the steps to analyze the choices: \\
1. Oil lamp. An oil lamp is designed to emit light using oil as fuel. Therefore, it does not fit the criteria as it indeed emits light. \\
2. LED lamp. An LED lamp uses light-emitting diodes to produce light. Like the oil lamp, it is designed to emit light, so it also does not fit the criteria. \\
3. Clamp. This option is a play on words. While "clamp" contains the word "lamp," it is not a type of lamp at all; instead, it's a tool used for holding objects tightly together. Since it's not a device designed to emit light, it technically "emits no light." \\
4. None of above. This option would be correct if none of the first three choices were accurate. However, based on the analysis, there is an option that meets the criteria of emitting no light in the context of the puzzle. \\
Given the play on words and focusing on the criteria of emitting no light, the correct option is: \\
3. Clamp. This is because it's the only choice among the options that, despite containing "lamp" in its spelling, does not function as a light-emitting device.
}

\end{document}